\documentclass[fleqn,10pt]{wlscirep}
\usepackage[utf8]{inputenc}
\usepackage[T1]{fontenc}
\usepackage{siunitx}
\usepackage[section]{placeins}

\newcommand{\figref}[1]{{Fig. \ref{#1}}}
\newcommand{\tabref}[1]{{Table \ref{#1}}}

\title{Exploring the Proprioceptive Potential of Joint Receptors Using a Biomimetic Robotic Joint}

\author[1,*]{Akihiro Miki}
\author[1]{Shun Hasegawa}
\author[1]{Sota Yuzaki}
\author[1]{Yuta Sahara}
\author[1]{Yoshimoto Ribayashi}
\author[1]{Kento Kawaharazuka}
\author[1]{Kei Okada}
\affil[1]{Mechano-Informatics, Graduate School of Information Science and Technology, the University of Tokyo, 7-3-1 Hongo, Bunkyo-ku, 1138656, Tokyo, Japan}

\affil[*]{miki@jsk.t.u-tokyo.ac.jp}


\begin{abstract}
  In neuroscience, joint receptors have traditionally been viewed as limit detectors, providing positional information only at extreme joint angles, while muscle spindles are considered the primary sensors of joint angle position.
  However, joint receptors are widely distributed throughout the joint capsule, and their full role in proprioception remains unclear.
  In this study, we specifically focused on mimicking Type I joint receptors, which respond to slow and sustained movements, and quantified their proprioceptive potential using a biomimetic joint developed with robotics technology.
  Results showed that Type I-like joint receptors alone enabled proprioceptive sensing with an average error of less than 2 degrees in both bending and twisting motions.
  These findings suggest that joint receptors may play a greater role in proprioception than previously recognized and that the relative contributions of muscle spindles and joint receptors are differentially weighted within neural networks during development and evolution.
  Furthermore, this work may prompt new discussions on the differential proprioceptive deficits observed between the elbows and knees in patients with hereditary sensory and autonomic neuropathy type III.
  Together, these findings highlight the potential of biomimetics-based robotic approaches for advancing interdisciplinary research bridging neuroscience, medicine, and robotics.
\end{abstract}
\begin{document}

\flushbottom
\maketitle
%
%
\thispagestyle{empty}


\section*{Introduction}\label{sec:joint_capsule_introduction}
\begin{figure}[htbp]
  \centering
  \includegraphics[width=0.5\textwidth]{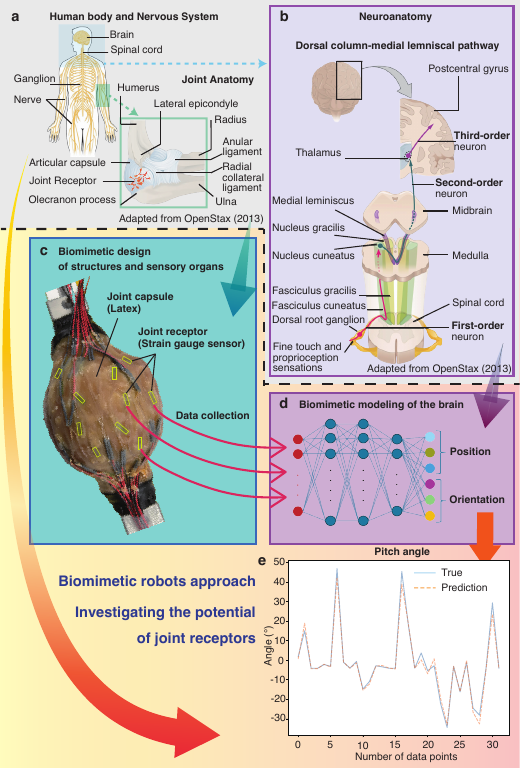}
  \caption{
    Overview of the conceptual framework and methodology of this study.
    This figure illustrates how joint receptors were anatomically and functionally modeled, implemented in a biomimetic joint with embedded strain gauge sensors, and evaluated using a neural network inspired by the dorsal column–medial lemniscal pathway.
    \textbf{a,} An anatomical diagram of the human body model and joint capsule is shown  (adapted from \cite{Betts:2013:CentralNervousSystem, Betts:2013:ElbowAnatomy}).
    \textbf{b,} The dorsal column–medial lemniscus pathway (DCML) is illustrated (adapted from \cite{Betts:2013:DCML}).
    \textbf{c,} Based on anatomical structures, the joint capsule and joint receptors are modeled and implemented on an actual device.
    \textbf{d,} The dorsal column–medial lemniscus pathway is modeled using a neural network, which is trained to output proprioceptive joint position sense from joint receptor data obtained from the actual device.
    \textbf{e,} The verification is conducted to determine whether proprioceptive joint position sense has been acquired through training.
  }
  \label{fig:joint-capsule-study-overview}
\end{figure}

{
  In neuroscience, proprioception—the unconscious perception of body position, movement, and muscular force output—has been extensively investigated.
  The role of joint receptors \cite{Wyke:1972:ArticularNeurology, lephart:1997:ProprioceptionRehabilitation, kandel:2000:NeuroScience, Riemann:2002:SensorimotorPart1}, located within the joint capsule \cite{ralphs1994joint}, has been a subject of research for decades.
  In the 1950s, joint receptors were thought to provide positional information across the full range of motion through collective discharge \cite{williams:1981:JointReceptorsRoll}.
  However, subsequent studies have challenged this assumption.
  Patients who underwent total hip replacement, which removes joint capsules and ligaments, retained both position and movement sense \cite{Grigg:1973:HipReplacement}.
  Similar findings were reported in cases of local knee joint anesthesia \cite{Clark:1979:KneePositionSense} and spinal lesions that blocked sensory input from the skin and joints without impairing lower limb movement sensation \cite{Wall:1977:TransectionDorsalColumns, Ross:1979:DorsalSpinocerebellarTracts}.
  Additionally, research on the thixotropic properties of muscles \cite{Proske:2012:ProprioceptiveReview} and proprioceptive illusions induced by tendon and muscle vibration \cite{Goodwin:1972:MuscleAfferentsKinaesthesia, Eklund:1972:PositionSenseVibration} further suggested that muscle spindles, rather than joint receptors, play the primary role in proprioception \cite{Proske:2011:KinestheticSenses, Proske:2012:ProprioceptiveReview}.
  Consequently, joint receptors have been considered mere limit detectors, providing positional feedback only at extreme joint angles.
  However, it has also been suggested that proprioception may involve not only muscle spindles \cite{Bewick:2015:MuscleSpinfleMechanoTransduction} and joint receptors,
  but also a variety of other sensory organs, such as Golgi tendon organs located in tendons \cite{Jami:1992:GolgiTendonOrgans} and cutaneous mechanoreceptors in the skin \cite{Johansson:1979:FourTypeTactileSensor, Johnson:2001:FourTypeMechanoreceptorsRoles}.
  Although multiple modalities are involved, the relative contributions of these sensory modalities and the mechanisms by which proprioceptive information is acquired remain unclear.
  To further advance our understanding, it is critical to quantitatively evaluate the latent capabilities of each sensory modality.
  In this study, we focus particularly on the potential of joint receptors.

  Recent neuroanatomical studies have continued to investigate joint receptors.
  For instance, in the shoulder joint, studies have shown that the density of joint receptors increases progressively toward the attachment regions of the glenoid and labrum \cite{witherspoon:2014:ShoulderSensoryInnervation}.
  A similar pattern has been observed in the elbow joint capsule, where receptor density in the mid-region, while lower than at the attachment regions, is still substantial, measuring approximately half of that found at the attachment regions \cite{Kholinne:2019:ElbowJointReceptorsDistribution}.
  If joint receptors only signal proximity to anatomical limits, they could be arranged locally to function as simple switches, and a widespread distribution, including in the mid-region of the joint capsule, would not be necessary.
  Furthermore, research has suggested that during pointing tasks, as the forearm approaches its extension limit, positional errors decrease, implying that populations of joint receptors may play a role in enhancing positional accuracy \cite{proske:2023:JointReceptorsReassessment}.
  Building upon these observations, we hypothesized that joint receptors may serve a broader function beyond merely acting as limit detectors at the extremes of joint motion.

  However, isolating the functional contributions of individual sensory receptors in vivo is challenging.
  Human neural networks develop throughout growth, and most studies assess subjects with mature sensorimotor integration.
  Additionally, proprioceptive feedback from joint receptors may be supplemented or compensated by other sensory modalities, making it difficult to quantify their intrinsic capabilities.
  Clinical findings also provide relevant background for this study.
  Hereditary Sensory and Autonomic Neuropathy Type III (HSAN III) is a congenital disorder characterized by the absence of muscle spindles \cite{Axelrod:2007:HSAN}.
  Patients with HSAN III exhibit marked ataxia but retain relatively preserved proprioception in the upper limbs, suggesting that, even in the absence of muscle spindle input, other sensory systems may contribute to maintaining proprioceptive function \cite{Macefield:2021:MechanoreceptorsProprioception, Macefield:2024:HSAN3ControlReview}.
  Given this background, the present study quantitatively investigates the proprioceptive potential of joint receptors using a biomimetic robotic approach.
  An overview of the study is shown in \figref{fig:joint-capsule-study-overview}.
  \figref{fig:joint-capsule-study-overview}a and b illustrate the distribution of joint receptors within the capsule and their connection to the central nervous system through the dorsal column–medial lemniscal pathway.
  Based on these biological insights, we focused on mimicking Type I joint receptors, which are widely distributed within the capsule and respond to slow and sustained movements.
  We constructed a soft biomimetic joint capsule made of latex rubber and embedded multiple strain gauge sensors to replicate the response characteristics of joint receptors (\figref{fig:joint-capsule-study-overview}c).
  The sensor signals, together with the measured joint angles and positions, were then used to train a neural network model inspired by the dorsal column–medial lemniscal pathway (\figref{fig:joint-capsule-study-overview}d).
  Finally, the model was evaluated to provide a foundation for further analysis and discussion (\figref{fig:joint-capsule-study-overview}e).

  This study adopts an empirical robotics-based approach to a problem that is difficult to address biologically, experimentally investigating whether proprioceptive functions can emerge solely from joint-receptor-like sensors.
  In contrast to conventional biomimetics research, which has primarily focused on the engineering reproduction of the structural and kinematic mechanisms of living organisms, the present study is positioned as a new constructive framework.
  This framework aims to reproduce the process through which integrated sensory functions emerge by imitating biological systems at a fundamental level of biological organization, specifically focusing on sensory receptors and their supporting connective tissues, and empirically examines the functional significance of these sensory organs.
  The objective of this study is to experimentally determine whether proprioceptive capability can emerge solely from joint receptor-like sensors and, through these findings, to provide both biological and engineering insights into the proprioceptive potential of joint receptors and the mechanisms of sensory integration.
}

\section*{Results} \label{sec:joint_capsule_results}

\subsection*{Biomimetic Joint with Joint Receptors Embedded in the Joint Capsule} \label{subsec:biomimetic_joint}
\begin{figure}[htbp]
  \centering
  \includegraphics[width=\linewidth]{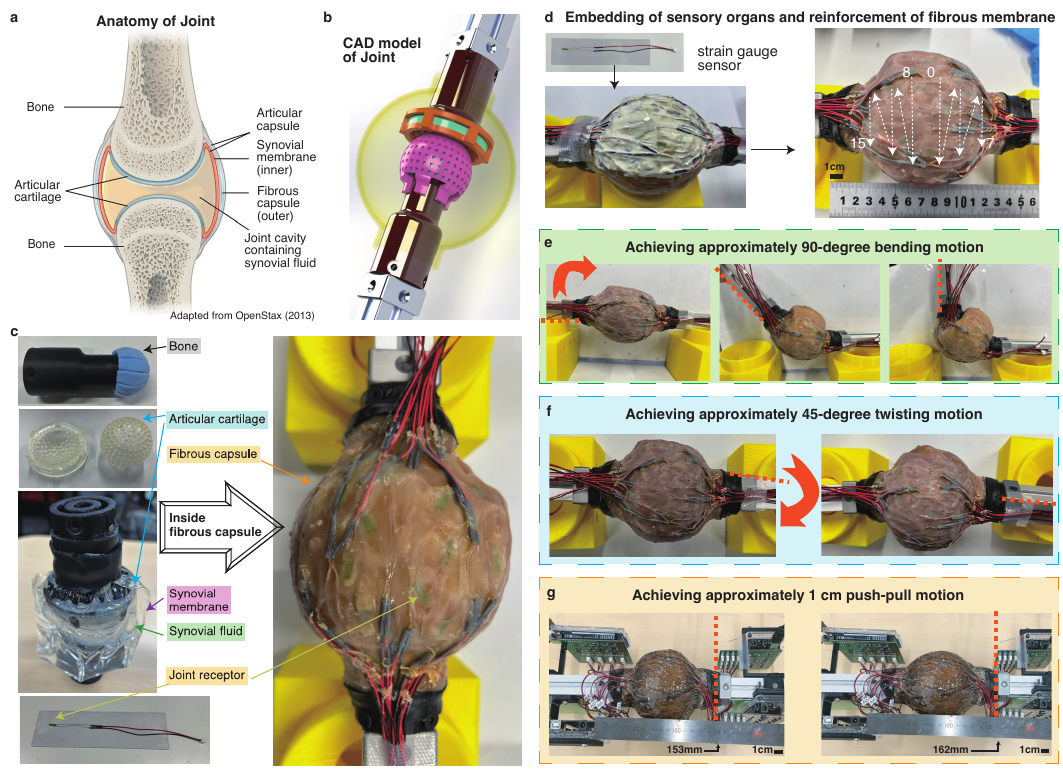}
  \caption{
    Design and implementation of the biomimetic joint used in this study.
    This figure illustrates how the biomimetic joint was designed with reference to anatomical structures, fabricated with embedded strain gauge sensors, and tested under different motion conditions (bending, twisting, and push–pull).
    \textbf{a,} Anatomical diagram of a biological joint. The joint capsule surrounds the joint formed by bone and cartilage. The anatomical figure is adapted from \cite{Betts:2013:OpenstaxSynovialJoints}.
    \textbf{b,} Design drawing of a biomimetic joint. The translucent yellow part represents the joint capsule.
    \textbf{c,} Various components of the biomimetic joint and the completed state after rubber molding. The 3D printed bone tips and cartilage parts are attached, and the soft joint capsule's fibrous membrane is created using natural rubber latex to enclose them.
    \textbf{d,} Diagram showing the process of shaping the joint capsule while embedding strain gauge sensors, which act as joint receptors. The sensors are arranged in three rows, with the higher-numbered sensors placed closer to the bone attachment points.
    \textbf{e,} Image showing the biomimetic joint bent at approximately 90 degrees.
    \textbf{f,} Image showing the biomimetic joint twisted at approximately 45 degrees.
    \textbf{g,} Image showing the biomimetic joint being pushed and pulled by approximately \SI{1}{\centi\meter}.
  }
  \label{fig:biomimetic-joint}
\end{figure}

{
  Since this study aims to investigate the potential of joint receptors, we developed a biomimetic joint in which joint receptors serve as fundamental components.
  Sensory receptors associated with joints are classified into four types based on their morphological characteristics \cite{Wyke:1972:ArticularNeurology}.
  Among these, Type 1 receptors have been reported as the most predominant in the elbow and shoulder joints \cite{Kholinne:2019:ElbowJointReceptorsDistribution, Kholinne:2021:Topography}.
  Accordingly, this study focuses on mimicking Type I receptors, which are known to respond to static pressure and exhibit a low adaptation rate.
  Their characteristics were reproduced using strain gauge sensors that change electrical output in response to static pressure and strain.
  Furthermore, the joint is designed at a life-size scale and incorporates soft material connections, allowing for joint motion with a scale and range comparable to those of the human body.
  By utilizing this biomimetic joint, we examine the potential of joint receptors in acquiring joint position sense.

  The joint was designed to replicate the anatomical structure depicted in \figref{fig:biomimetic-joint}a and is shown in \figref{fig:biomimetic-joint}b.
  The distal end of the aluminum frame is equipped with 3D-printed bone ends and cartilage components, while a TPU film mimics the synovial membrane by enclosing a pseudo-synovial fluid.
  Additionally, the soft fibrous membrane of the joint capsule is created using natural rubber latex (\figref{fig:biomimetic-joint}c).
  During the joint capsule molding process, strain gauge sensors representing joint receptors were embedded as extensively as possible (\figref{fig:biomimetic-joint}d).
  The thickness of the joint capsule is approximately \SI{2}{\milli\meter}, and the sensors were arranged in three rows, numbered from the midsection of the joint capsule toward the bone attachment side.
  As a result, a total of 60 sensory receptors were embedded within a spherical joint with a diameter of approximately \SI{100}{\milli\meter}.

  Figure panels \figref{fig:biomimetic-joint}e, f, and g illustrate the movement of the fabricated biomimetic joint.
  As designed, the joint achieved an approximately \SI{90}{\degree} range of motion in all directions for bending movements (\figref{fig:biomimetic-joint}e).
  For twisting (axial rotation; pronation–supination) movements, the joint exhibited an approximately \SI{45}{\degree} range of motion along the axial direction when fully extended, comparable to that of the human elbow joint (\figref{fig:biomimetic-joint}f).
  In addition, since it is an open-type ball joint, push–pull movements were also possible, achieving approximately \SI{1}{\centi\meter} of displacement (\figref{fig:biomimetic-joint}g).
  These flexible movements can be attributed to the softness of the fibrous latex membrane, indicating that the joint successfully reproduces the compliant mechanical behavior of biological joints.
}

\subsubsection*{Performance Verification of Joint Receptors Using Deep Learning} \label{subsec:experiments_estimate_proprioception}
\begin{figure*}[htbp]
    \centering
    \includegraphics[width=\linewidth]{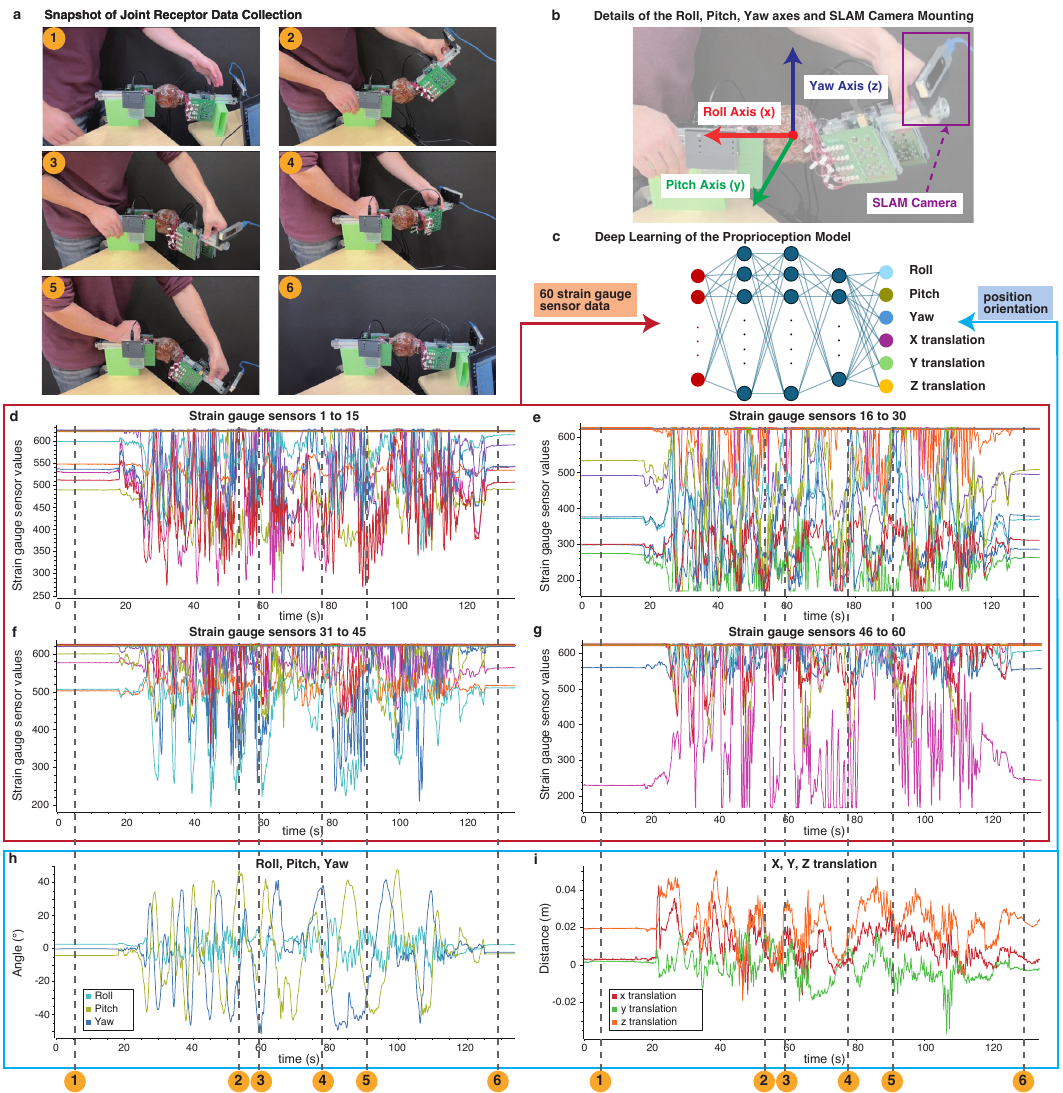}
    \caption{
      Data acquisition and modeling framework for proprioception estimation using the biomimetic joint.
      \textbf{a,} The process of moving the biomimetic joint to acquire data from 60 strain gauge sensors embedded in the joint capsule, functioning as joint receptors, and joint position and orientation data.
      \textbf{b,} Joint movement data is obtained using SLAM Camera (Intel RealSense T265), providing relative coordinates and orientations from the initial posture. The mounting position of the camera and its coordinate system are shown.
      \textbf{c,} A deep learning model was developed to estimate the joint's position and orientation from receptor data. The input consists of values from the 60 strain gauge sensors, and the output includes roll, pitch, yaw, and x, y, z translations.
      \textbf{d-i,} Visualization of changes in sensor values during movement (d–g), and corresponding orientation (h) and position (i).
      Sensor values are shown only for connected sensors and were read via a voltage divider circuit.
      The horizontal axis of all graphs represents time in seconds.
      The vertical axis of the sensor graphs corresponds to voltage readings from the circuit (with the top value of approximately 628 corresponding to 3.3 V), while those of the roll, pitch, and yaw graphs represent angles, and those of the x, y, and z graphs represent positions in meters.
      Annotations 1–6 correspond to panel (a).
    }
    \label{fig:data-collection-for-joint-receptors}
  \end{figure*}

\begin{figure*}[htbp]
    \centering
    \includegraphics[width=\linewidth]{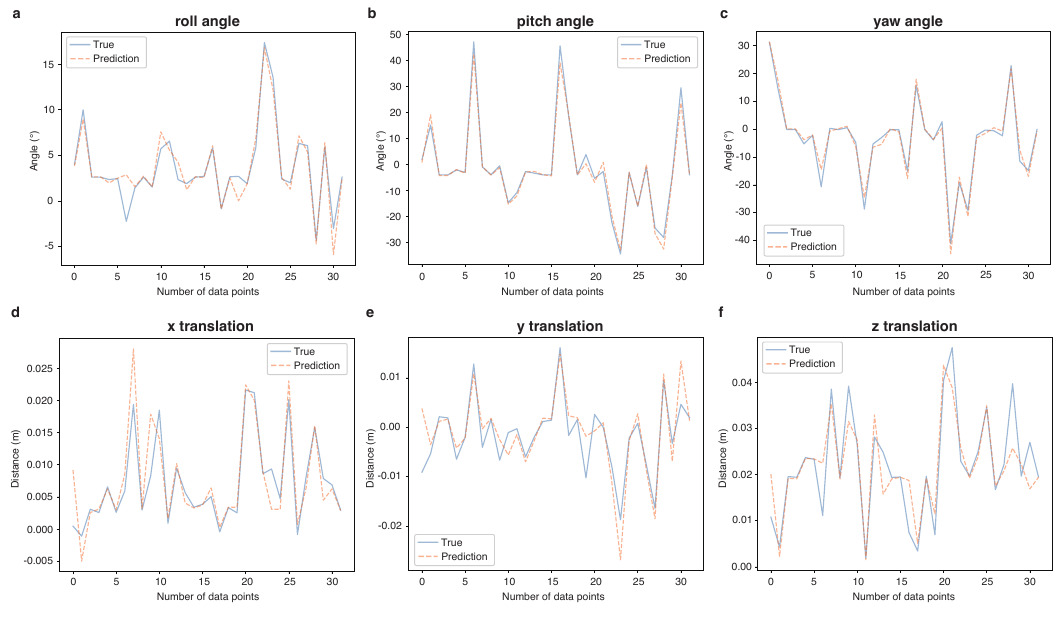}
    \caption{
      Performance of proprioception estimation using the trained deep learning model.
      \textbf{a-f,} This graph shows the estimation of the joint's coordinates and orientation generated using deep learning based on sensory receptor data collected from 60 strain gauge sensors embedded in the joint capsule.
      Due to the open-type ball joint loosely connected by soft tissues, the graph visualizes not only orientation(roll(a), pitch(b), and yaw(c)) but also translation(x(d), y(e), and z(f) positions), comparing them to actual measurements (ground truth).
      The vertical axis represents angles for roll, pitch, and yaw as well as position in meters, while the horizontal axis represents a dimensionless number denoting the number of test data points.
    }
    \label{fig:joint-receptors-performance}
  \end{figure*}

{
  An experimental validation was conducted using a deep learning model inspired by the dorsal column–medial lemniscal pathway to determine whether joint angles and positions could be estimated as proprioceptive information from a large number of strain gauge sensors embedded in the joint capsule, functioning as joint receptors.
  An overview of the data acquisition and learning process is shown in \figref{fig:data-collection-for-joint-receptors}.
  The relative position and orientation of the joint from its initial posture, representing proprioception, were obtained using Visual SLAM \cite{Fuentes:2015:VisualSLAM} with the Intel RealSense T265 \cite{Schmidt:2019:RealsenseT265}.
  \figref{fig:data-collection-for-joint-receptors}b shows the attachment of the Intel RealSense T265 and the roll, pitch, and yaw axes.
  The relationships between the sensor responses and the joint movements are presented in \figref{fig:data-collection-for-joint-receptors}d–i.
  A total of 1263 state values were collected.

  The performance of the trained deep learning model in proprioception estimation, compared with actual measured values, is illustrated in \figref{fig:joint-receptors-performance}a-f.
  The open-type ball joint, connected via soft tissues, permits not only twisting and bending movements but also push-pull displacements.
  Consequently, the six estimated outputs include not only roll, pitch, and yaw orientations but also translations corresponding to push-pull movements.
  Additionally, \tabref{tab:proprioception-error-statistics-xyz} summarizes statistical error metrics, including maximum error, mean error, and standard deviation for position estimation, whereas \tabref{tab:proprioception-error-statistics-rpy} provides the same metrics for orientation estimation.
  Here, the mean error is defined as the average of the absolute error values (|predicted – actual|) across all samples, and the standard deviation is also computed from the absolute errors.

  As a result, even when using only Type I–like joint receptors, the model successfully estimated joint angles during bending and twisting motions with a mean error within \SI{2}{\degree}.
  Although the acquired data contained a small offset (see Methods: Data Collection Details for specifics), the deep learning model achieved accurate proprioceptive estimation, remaining within a mean error of \SI{2}{\degree} under these conditions.
  This confirms that the influence of such minor offsets on the results was negligible.
}

  \begin{table}[htbp]
    \centering
    \caption{Statistical Values on Positional Errors.}
    \label{tab:proprioception-error-statistics-xyz}
    \small
    \begin{tabular}{|c|c|c|c|}
      \hline
      Error Type & x (m) & y (m) & z (m) \\
      \hline
      Maximum Error   & 0.0195 & 0.0149 & 0.0247 \\
      Mean Error      & 0.0023 & 0.0020 & 0.0033 \\
      Standard Deviation & 0.0028 & 0.0022 & 0.0037 \\
      \hline
    \end{tabular}
  \end{table}

  \begin{table}[htbp]
    \centering
    \caption{Statistical Values on Attitude Errors.}
    \label{tab:proprioception-error-statistics-rpy}
    \small
    \begin{tabular}{|c|c|c|c|}
      \hline
      Error Type & roll (°) & pitch (°) & yaw (°) \\
      \hline
      Maximum Error   & 6.1621 & 7.8585 & 9.4371 \\
      Mean Error      & 0.7211 & 1.3025 & 1.3744 \\
      Standard Deviation & 0.8798 & 1.4239 & 1.5882 \\
      \hline
    \end{tabular}
  \end{table}

\subsubsection*{Analysis of Redundancy in Joint Receptors} \label{subsec:experiments_verification_of_redundancy}
\begin{figure}[htbp]
    \centering
    \includegraphics[width=\linewidth]{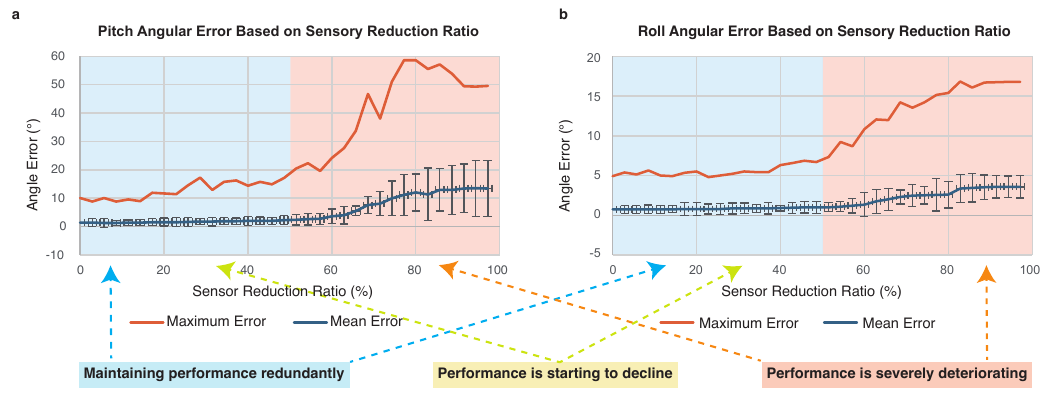}
    \caption{
      Redundancy analysis of proprioception estimation under reduced sensor input with ten independent trials.
      This figure shows the mean and standard deviation of angular errors across ten independent trials for different sensor reduction ratios.
      Panels (a) and (b) correspond to pitch and roll, respectively.
      Error bars represent the standard deviation across trials for each reduction ratio, magnified tenfold for visibility.
      Statistical analysis revealed that mean angular error first showed a statistically significant increase compared with the baseline at a reduction ratio of \SI{25.7}{\percent} for pitch and \SI{28.6}{\percent} for roll ($p < 0.05$).
      In contrast, maximum angular error exhibited statistically significant increases only at higher reduction ratios (\SI{48.6}{\percent} for pitch and \SI{51.4}{\percent} for roll).
      These results indicate that while average performance degradation begins around a \SI{30}{\percent} reduction, extreme errors become prominent as the reduction approaches \SI{50}{\percent}.
      Notably, a mean angular error exceeding \SI{2}{\degree} was first observed at a reduction ratio of \SI{45.7}{\percent} in pitch.
    }
    \label{fig:redundancy-of-joint-receptors}
  \end{figure}

{
  Humans possess a redundant sensory system that allows them to retain proprioception even in the presence of minor injuries or damage.
  In this study, we conducted an experiment to determine whether the sensory system of the biomimetic joint developed in this work also exhibits redundancy due to the presence of multiple sensory receptors.
  By progressively reducing the number of sensory receptors, we investigated the impact on the performance of a deep learning model tasked with estimating proprioception from joint receptor data.
  The details of the analysis procedure are provided in the Methods section (Methods for Analyzing Redundancy and Receptor Importance in the Biomimetic Joint).

  The results of the redundancy analysis are shown in \figref{fig:redundancy-of-joint-receptors}.
  The horizontal axis represents the sensor reduction ratio relative to the total number of sensors, and the vertical axis represents the joint angle error.
  For the pitch and roll axes, which correspond to large bending and twisting movements, the mean and maximum errors across ten independent trials are shown.
  Error bars indicate the standard deviation across trials, magnified tenfold for visibility.
  Statistical analysis (Welch’s t-test with Holm correction, baseline: \SI{0}{\percent} condition) revealed that a statistically significant increase in mean angular error was observed at a reduction ratio of \SI{25.7}{\percent} for pitch and \SI{28.6}{\percent} for roll ($p < 0.05$).
  In contrast, maximum angular error exhibited statistically significant increases at higher reduction ratios, specifically at \SI{48.6}{\percent} for pitch and \SI{51.4}{\percent} for roll.
  Furthermore, a mean angular error exceeding \SI{2}{\degree} was first observed at \SI{45.7}{\percent} reduction in pitch.
  These findings indicate that average performance degradation begins around a reduction ratio of \SI{30}{\percent}, whereas extreme errors become prominent near \SI{50}{\percent}.
  In addition, considering that 20 of the 60 joint receptors had already failed at the start of the experiment, only 40 receptors were functional.
  At a reduction ratio of \SI{50}{\percent}, this corresponds to only 20 functional receptors (40 $\times$ (1 – 0.5)), yet proprioceptive function was still partially maintained up to that point.
  This demonstrates that the biomimetic joint developed in this study possesses redundancy in proprioception, retaining functionality even after losing nearly half of its joint receptors.
  A representative result from a single trial is shown in Supplementary \figref{fig:redundancy-of-joint-receptors-sample}.
  Since this is a single trial, statistical analysis is not applicable; however, visual inspection suggests a clear degradation trend emerging near a reduction ratio of \SI{50}{\percent}, which is consistent with the averaged results in the main text.
  In \figref{fig:redundancy-of-joint-receptors}, error bars indicate the standard deviation of errors across trials at each reduction ratio.
  For better visibility, these values were magnified tenfold. In contrast, Supplementary \figref{fig:redundancy-of-joint-receptors-sample} shows a representative result from a single trial, where error bars correspond to the within-trial standard deviation without scaling.
}

\subsubsection*{Distribution of Critical Sensory Receptors Near Range of Motion Limits} \label{subsec:experiments_verification_of_distribution}
\begin{figure}[htbp]
    \centering
    \includegraphics[width=\linewidth]{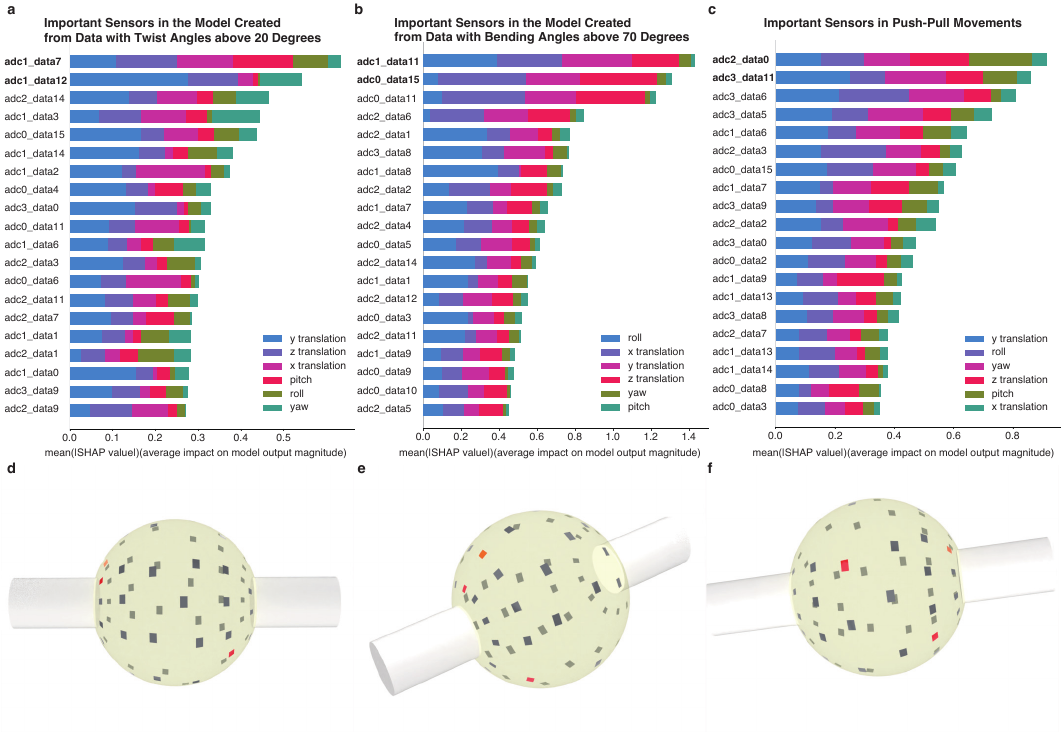}
    \caption{
      Identification of key joint receptors for proprioception estimation.
      This figure highlights the sensory receptors that the deep learning model identified as most important during different joint movements, together with visualizations of their spatial distribution.
      Graphs showing key joint receptors important for the deep learning model during twisting (a), bending (b), and push-pull (c) movements near the limit of the joint's range of motion, along with their respective visualization renderings (d corresponds to a, e to b, and f to c).
      Each receptor was labeled as “adc(i)\_data(j),” where i corresponds to the board number and j to the sensor number, arranged in three rows from mid-capsular region as shown in \figref{fig:biomimetic-joint}d.
      Color coding reflects both the frequency and magnitude of receptor importance across the five model-training trials.
      The receptor that appeared most frequently among the top three across trials is shown in red, the remaining one or two receptors within the top three are shaded according to their SHAP values, and all other receptors are displayed in dark gray.
    }
    \label{fig:joint-receptors-comparison-to-human}
  \end{figure}

{
  As noted in the Introduction, recent neuroscientific perspectives suggest that joint receptors in the human body play a significant role near the limits of joint angle range.
  Moreover, previous studies have reported that the density of joint receptors is higher near bone attachment regions than in the midsection of the joint capsule \cite{witherspoon:2014:ShoulderSensoryInnervation, Kholinne:2019:ElbowJointReceptorsDistribution}, suggesting that receptors near bone attachment regions may play a particularly important role.
  In this study, we examined the spatial distribution of key sensory receptors in the biomimetic joint we developed, with a particular focus on regions near the limits of its range of motion during twisting, bending, and axial (push-pull) movements.
  By evaluating whether these critical receptors tend to be concentrated near the bone attachment regions—similar to the distribution observed in human joints—we aimed to assess the anatomical and functional similarity between the artificial and biological joints.

  For twisting and bending movements, additional data were collected using the method outlined in the Performance Verification of Joint Receptors Using Deep Learning section.
  From this dataset, instances where the joint angles exceeded a predefined threshold were extracted.
  A deep learning model was then trained to estimate proprioception from joint receptor data, as described in the Performance Verification of Joint Receptors Using Deep Learning section.
  The importance of each joint receptor was assessed to analyze the distribution of critical sensory receptors near the limits of joint motion (details of the analytical method are provided in Methods, Methods for Analyzing Redundancy and Receptor Importance in the Biomimetic Joint).
  Since deep learning models introduce stochasticity, training the same model multiple times may yield different results.
  To address this, we trained five independent models for each movement type and counted how many of the top three receptors (per trial) belonged to the mid-capsular, transitional, or bone attachment regions.
  In twisting movements, the counts of receptors located in the mid-capsular, transitional, and bone attachment regions were 3, 6, and 6 out of 15, respectively, indicating that influential receptors were predominantly situated near the bone attachment or within the transitional region.
  Furthermore, adc1\_data7 and adc1\_data12 appeared within the top three in all five trials, and these receptors correspond to the bone attachment and transitional regions, respectively.
  In bending movements, the counts were 0, 9, and 6 out of 15 for the mid-capsular, transitional, and bone attachment regions, again showing that the most important receptors were concentrated near the bone attachment or its transitional region.
  In addition, adc0\_data15 and adc1\_data11 consistently appeared among the top three across all trials, and these receptors are located in the bone attachment and transitional regions, respectively.
  An example evaluation result is shown in \figref{fig:joint-receptors-comparison-to-human}a, b, visualizing the top three receptors (\figref{fig:joint-receptors-comparison-to-human}d, e).
  As illustrated in the bar graphs in \figref{fig:joint-receptors-comparison-to-human}, importance values were skewed toward the top-ranked receptors, while lower-ranked receptors exhibited similar importance scores.
  These findings indicate that, for both twisting and bending near the joint’s range limits, important receptors were predominantly located around the bone attachment region or its transitional region.
  Considering that human joint receptors are also densely distributed near bone attachment regions, these results highlight a similarity between the biomimetic and biological joints.

  Next, pushing and pulling movements were analyzed using the same data collection method as in the Performance Verification of Joint Receptors Using Deep Learning section, ensuring that only movements strictly aligned with the bone direction were considered.
  Since the range of motion for pushing and pulling in the developed joint was approximately \SI{1}{\centi\meter}, we extracted data where the bone contact distance deviated by more than \SI{2}{\milli\meter} from the initial position (either pushed or pulled).
  A deep learning model was trained, and joint receptor importance was evaluated.
  The results are presented in \figref{fig:joint-receptors-comparison-to-human}c, f.
  For push–pull movements, five models were likewise trained, and the locations of the top three receptors were counted across the mid-capsular, transitional, and bone attachment regions.
  The counts were 5, 6, and 4 out of 15, respectively, indicating that, unlike twisting and bending, a relatively large number of influential receptors were also found in the mid-capsular region.
  In addition to adc3\_data11, which is located in the transitional region, adc2\_data0 frequently appeared among the top three across trials.
  This receptor belongs to the mid-capsular region, and its repeated selection suggests that receptors in the mid-capsular region, not only those near the bone attachment, contribute substantially to proprioceptive estimation during push–pull movements.
}

\section*{Discussion} \label{sec:joint_capsule_discussion}
{
  Neuroscience seeks to understand how the nervous system's networks in the human body develop and adapt.
  Among these, proprioception plays a crucial role in posture and movement control through the integration of information from multiple sensory organs, including muscle spindles, cutaneous mechanoreceptors, and joint receptors.
  To explore how these receptors acquire proprioception, it is crucial to assess the information each receptor type provides.
  In this study, we addressed this neuroscientific challenge, which is difficult to investigate directly in biological systems, by taking a constructive approach that empirically explores how integrated sensory functions can emerge through the imitation of sensory receptors and their supporting tissues.
  Following this approach, we developed a biomimetic joint and experimentally examined the potential of joint receptors to acquire proprioceptive capability.

  This section outlines the constraints of the biomimetic joint used in this study.
  The joint capsule was made of isotropic elastic latex, whereas human joint capsules contain anisotropic collagen fibers \cite{Ng:2019:HipJointCapsule}.
  To develop more advanced joints, future work may involve refining rubber molding, bioengineering new materials, or designing anisotropic structures.
  Recent advances in artificial skin culturing \cite{Kawai:2022:LivingSkin} and anisotropic metamaterials \cite{Zhang:2022:AnisotropyMaterial, Bodaghi:2017:SoftMetamaterials, Mirzaali:2018:MultiMaterialMetamaterials} provide promising directions.
  Nevertheless, as confirmed in the Biomimetic Joint with Joint Receptors Embedded in the Joint Capsule section, the biomimetic joint achieved 90-degree bending, 45-degree twisting, and \SI{1}{\centi\meter} push-pull motions.
  The soft-tissue-connected open ball joint allowed flexible movement, approximating human joint motion.
  Given the human elbow's range of motion, this implementation is sufficiently soft and functional within its intended scope.
  There are unavoidable differences between the biomimetic and human joint capsules regarding sensory receptors.
  First, we compare receptor number and density.
  Human elbow joint capsules contain 1–2 sensory receptors per \SI{1}{\cubic\centi\meter} \cite{Kholinne:2019:ElbowJointReceptorsDistribution}.
  The biomimetic capsule, with \SI{2}{\milli\meter} latex layered on a \SI{10}{\centi\meter} hollow sphere, had a volume of \SI{62.84}{\cubic\centi\meter}.
  With 60 embedded receptors, its density was approximately one per \SI{1}{\cubic\centi\meter}, similar to human values.
  In addition to number and density, the spatial arrangement of receptors also differs between the biomimetic and human joint capsules.
  In humans, sensory receptors tend to be more densely distributed near bone attachment regions, although they are also present throughout the capsule \cite{Kholinne:2019:ElbowJointReceptorsDistribution}.
  In the biomimetic capsule, however, due to technical constraints such as sensor size and embedding methods, it was not feasible to replicate this anatomical distribution.
  Instead, sensors were uniformly distributed, which resulted in a three-row configuration.
  Although this arrangement does not fully match the anatomical distribution in the human joint capsule, our experiments demonstrated that it was sufficient to reconstruct joint angles with reasonable accuracy.
  Moreover, the analysis of this simplified arrangement allowed us to investigate trends related to the placement of important sensors.
  Nevertheless, differences from the human anatomical structure may still influence the results, and this limitation should be taken into account when interpreting the findings.
  Regarding receptor types, this study focuses exclusively on Type 1 receptors, which respond to slow movements.
  Type 2 receptors, sensitive to rapid motion, were not considered. Future research should address this limitation to examine rapid motion responses.
  The size of the joint receptors also differs significantly.
  Human joint receptors are typically on the order of \SI{100}{\micro\meter} \cite{Kholinne:2019:ElbowJointReceptorsDistribution, Wyke:1972:ArticularNeurology}, whereas current sensor technology makes it extremely challenging to replicate this scale.
  In this study, strain gauge sensors were used as substitutes, though their significantly larger size may have influenced the results.
  The representation of sensory information differs between robots and humans.
  Robots relay numerical data to a central computer, whereas humans transmit firing frequencies via synapses.
  However, since this study focuses on the functional capabilities of joint receptors, the differences in representation format are not expected to significantly impact the quality of decoded information.
  Consequently, the representation format was not a target for biomimicry.
  For the same reason, unique characteristics of human nerve fibers, such as adaptation and variations in transmission speed, were also not replicated in this study.

  Based on the above considerations, as shown in the Performance Verification of Joint Receptors Using Deep Learning section, this study achieved proprioception with a large number of joint receptors, maintaining an average error within 2 degrees for the roll, pitch, and yaw axes.
  Previous studies \cite{Goble:2010:ProprioceptiveAssessment, Macefield:2016:HSAN3KneeTapingProprioception, Smith:2020:HSAN3ElbowTapingProprioception} have reported that human proprioception exhibits an average error of approximately 2 to 3 degrees.
  Thus, the experimental results demonstrate a level of accuracy comparable to biological systems, supporting the effectiveness of the proposed biomimetic approach.
  Several factors may account for the observed prediction errors.
  One possibility is the uncertainty regarding whether biological sensory systems apply fine-tuned corrections to individual sensory receptors.
  Thus, we did not apply hysteresis correction to each sensor but evaluated overall behavior.
  Furthermore, the deep learning model used for proprioception estimation in this study employed a simple architecture to examine the feasibility of proprioception estimation without assuming complex neural processing.
  While more advanced architectures, such as LSTM \cite{Hochreiter:1997:LSTM} and Transformer \cite{Vaswani:2017:Transformer}, have been widely applied to biological data \cite{Yildirim:2019:ArrhythmiaLSTMClassification, Sun:2021:EEGTransformer}, evaluating the potential improvement in accuracy with such methods remains an area for future research.
  Nevertheless, the results demonstrate that accurate proprioception can be achieved without incorporating complex models.

  This study also compared the biomimetic joint to the human body in terms of joint receptor redundancy and the distribution of critical sensory receptors near joint limits.
  %
  For bending and twisting movements, statistical analysis of angular errors in pitch and roll showed that a statistically significant increase was observed at around a \SI{30}{\percent} reduction ratio.
  However, if the inherent ambiguity of human proprioception is assumed to be approximately 2 degrees, this threshold was first exceeded at \SI{45.7}{\percent} reduction.
  Furthermore, when focusing on maximum error, which is more critical for joint function, statistically significant increases were observed near \SI{50}{\percent}.
  These results indicate that proprioceptive function can be maintained until nearly half of the joint receptors are lost, confirming redundancy in the system.
  Although the exact proportion of receptor loss that humans can tolerate remains unclear, it is widely recognized that minor injuries or receptor damage do not result in a complete loss of proprioception.
  Our findings in the Analysis of Redundancy in Joint Receptors section suggest the biomimetic joint exhibits similar redundancy to humans.
  %
  In analyzing the distribution similarity of critical sensory receptors near joint angle limits the Distribution of Critical Sensory Receptors Near Range of Motion Limits section, the results for bending (pitch) and twisting (roll) movements indicated that receptors located near the bone attachment regions or within the transitional regions exhibited higher importance.
  Considering that human joints show higher receptor density near bone attachment regions, these findings suggest that the biomimetic joint replicates human-like patterns of important receptor distribution near the limits of joint motion.
  For push-pull movements, however, receptors in the mid-capsular regions exhibited higher importance than those near bone attachment regions.
  While this contrasts with human receptor distribution, it may stem from this study's focus solely on joint capsule receptors, omitting ligaments, bones, and cartilage, which respond strongly in push-pull movements.
  Future research should aim to construct models that incorporate sensory receptors across various tissues.

  The confirmation of redundancy across joint receptors and the similarity in critical receptor distributions near joint angle limits indicate that the biomimetic joint developed in this study serves as a valid approximation of its biological counterpart.
  Accordingly, the finding that proprioceptive information could be derived solely from the joint-receptor-like sensors suggests the possibility that humans may also acquire proprioceptive cues primarily through the integration of joint receptor signals.
  In biology and neuroscience, proprioception is typically examined in living organisms, where sensory responses are integrated within a fully developed and adaptive nervous system.
  Consequently, measurements of proprioceptive function often reflect responses that have already undergone developmental adaptation and multisensory integration, making it difficult to isolate the contributions of individual sensory receptors.
  Moreover, the information obtained from joint receptors in such biological contexts may be partially compensated for or complemented by inputs from other sensory modalities.
  For these reasons, assessing the intrinsic potential of individual receptors directly within biological systems remains a major experimental challenge.
  In this study, we adopted a constructive framework that reproduces joint receptors and their supporting tissues, thereby enabling the investigation of sensory emergence on a platform free from the complexities of biological neural development and multisensory compensation.
  This framework offers an empirical basis for examining the inherent potential of joint receptors through the observation of emergent proprioceptive functions.

  According to prevailing views in neuroscience, joint receptors primarily function as limit detectors, while muscle spindles are considered the main sensors for proprioception.
  If joint receptors do possess proprioceptive potential, it is plausible that, during development or evolution, the human neural network has shifted weight toward muscle spindles for proprioceptive dominance.
  The findings of this study prompt interdisciplinary discussion across robotics, biology, and neuroscience regarding how such neural weighting may emerge, providing a foundation for cross-disciplinary understanding.

  In relation to this discussion, a condition described in the Introduction is Hereditary Sensory and Autonomic Neuropathy Type III (HSAN III).
  HSAN III has been studied neurophysiologically as a congenital model in which muscle spindles are absent \cite{Macefield:2011:HSAN3MuscleSpindleLoss, Macefield:2013:HSAN3KneeProprioception, Smith:2018:HSAN3HandMuscleSpindle}.
  These studies have provided insights into the relative roles of muscle spindles and cutaneous mechanoreceptors \cite{Macefield:2016:HSAN3KneeTapingProprioception, Smith:2020:HSAN3ElbowTapingProprioception, Macefield:2021:MechanoreceptorsProprioception, Macefield:2024:HSAN3ControlReview}.
  However, the role of joint receptors has not been thoroughly investigated.
  If the latent functional potential of joint receptors is considered, it is plausible that their contribution may expand in HSAN III, where muscle spindles are absent.
  Further investigation of joint receptors in HSAN III patients could provide new insights into the neural organization and adaptive formation of proprioceptive networks.
  Interestingly, in HSAN III, proprioception in the knee joint is markedly impaired, whereas proprioception in the elbow joint remains nearly normal \cite{Macefield:2016:HSAN3KneeTapingProprioception, Smith:2020:HSAN3ElbowTapingProprioception}.
  Smith et al. \cite{Smith:2020:HSAN3ElbowTapingProprioception} described this discrepancy as puzzling, but joint receptors may offer a possible explanation.
  A major difference between the knee and the elbow is that the knee is subjected to greater and more frequent loads and impacts due to locomotion and daily activities.
  Joint receptors are known to contribute to joint stability \cite{Lundberg:1978:JointAfferentsMotorControl, Zimny:1991:KneeJointNeuroreceptors, Riemann:2002:SensorimotorPart2}, and they are likely to respond to mechanical shocks.
  If joint receptors compensate for the loss of muscle spindles by taking on proprioceptive functions, they may also be involved in stabilizing the joint against impacts.
  This dual demand could lead to a redistribution of functional weighting within the proprioceptive neural network.
  In this case, the knee joint receptors, burdened by both stability and proprioception, may contribute less specifically to position sense.
  Such redistribution may underlie the pronounced deterioration of proprioceptive performance observed in the knee joint of HSAN III patients.

  The constructive approach adopted in this study differs from conventional biomimetics research, which has primarily focused on the engineering reproduction of structural mechanisms and motor functions in living organisms.
  In contrast, by mimicking sensory receptors and their supporting tissues at a finer level of biological organization, this study introduces a new framework for reconstructively examining how integrated sensory functions can emerge.
  This framework provides an experimental foundation for generating new discussions and hypotheses, such as those regarding the potential of joint receptors and their possible involvement in HSAN III.
  Such an empirical and constructive methodology offers a new experimental paradigm for understanding biological functions through the integration of neuroscience and robotics.
  Future research could expand on this study, which focused on joint capsules and receptors, by incorporating other sensory components, such as the skin around the joints and the muscles that drive movement.
  By utilizing robotic systems, we can explore neural network development in response to complex stimuli, such as impacts and contact, potentially advancing discussions in biology and neuroscience, including those related to HSAN III.
  The sensory systems of the skin and muscles, together with the joint system, are key components of the somatosensory system.
  Integrating these systems could enable a more comprehensive understanding of somatosensory function.
  Additionally, robotics offers the unique advantage of conducting quantitative analysis on individual muscles and sensory units, allowing for applications in fields like sports medicine through load analysis of specific movements.
  This research has significant potential for advancing interdisciplinary fields, including neuroscience, medicine, biology, and robotics.
}


\section*{Methods} \label{sec:methods}

\subsection*{Details of the Biomimetic Joint Construction} \label{subsec:details_biomimetic_joint_construction}
{
  Although the constructed joint exhibits motion characteristics similar to those of the human elbow joint, it was intentionally designed as a generalized synovial ball joint to investigate the fundamental potential of joint receptors without being constrained by any specific anatomical structure.
  Because accurately simulating soft tissues such as the joint capsule is challenging and often inaccurate, we constructed a physical biomimetic joint to experimentally investigate the potential function of joint receptors.
  The biomimetic joint was designed using SolidWorks 2020 (Dassault Systèmes, Waltham, MA, USA), a mechanical design software.
  This section provides a detailed description of the various components that constitute the biomimetic joint.
  The external bone structure of the joint was fabricated using aluminum frames from MISUMI Group Inc. (Tokyo, Japan).
  The internal bone, which comes into contact with joint fluid, was constructed from a filament composed of polyamide 12 (PAHT-CF, Bambu Lab Inc., Shenzhen, China) mixed with carbon fiber.
  This composite material offers both strength and water absorption properties, making it ideal for the bone structure.
  A Bambu Lab X1-Carbon 3D printer (Bambu Lab Inc., Shenzhen, China) was used for the printing process.
  The cartilage component, which attaches to the tip of the bone, was created using a stereolithography (SLA) 3D printer (Form3, Formlabs Co., Somerville, MA, USA) and an elastic rubber-like resin (Elastic50A, Formlabs Co., Somerville, MA, USA).
  The cartilage was then affixed to the bone at the joint connection using adhesive.
  These components were encapsulated in a polyurethane film (Touch Grace Film, Takeda Sangyo Co., Tokyo, Japan) to replicate the synovial membrane of the joint, along with joint fluid, which consisted of a \SI{3}{\milli\gram\per\milli\liter} hyaluronic acid solution designed to mimic human synovial fluid.
  To simulate the fibrous membrane of the joint capsule, the outer layer was made from natural rubber latex (PC-518, Regitex Co., Ltd., Kanagawa, Japan).
  The final thickness of the fibrous membrane was approximately \SI{2}{\milli\meter}, achieved through multiple layers of application and drying.
  Strain gauge sensors (General-purpose Foil Strain Gages, KYOWA ELECTRONIC INSTRUMENTS CO., LTD, Tokyo, Japan) were embedded within the fibrous membrane during fabrication to serve as joint receptors.
  The strain gauge sensors were manually embedded into the rubber-based capsule, resulting in approximate spacing but with variations in both spacing and orientation due to manual placement.
  The connection between the joint capsule and the bone was secured using heat shrink tubing (TREDUX-MA47, HellermannTyton Co. Ltd - Japan, Tokyo, Japan).
}

\subsection*{Strain Gauge Sensor Reading Circuit Board} \label{subsec:strain_gauge_sensor_reading_circuit_board}
{
  The circuit board for reading strain gauge sensors, which function as joint receptors, was designed using KiCad (Version 7.0, available at https://www.kicad.org/).
  A Wheatstone bridge configuration, consisting of resistors, was used for each strain gauge sensor to form a voltage divider circuit.
  To amplify the signal, an instrumentation amplifier (INA2126U, Texas Instruments Inc.) was employed, providing a gain of 1000.
  The amplified signal was then read by an Arduino MEGA \cite{Arduino:2015:Arduino, Banzi:2022:GettingStartedArduino}.
  Due to the voltage divider circuit, the strain gauge sensor outputs were within the range of \SI{0}{\volt} to \SI{3.3}{\volt}.
  The value of approximately 628 at the peak of the graph in \figref{fig:data-collection-for-joint-receptors}d-g corresponds to \SI{3.3}{\volt}.
  The changes in the graph during disconnection are described in \figref{fig:supplementary-wire-disconnecting-plot}.
  A total of 15 strain gauge sensors were interfaced with a single Arduino MEGA, with four units utilized to achieve the required readings.
}

\subsection*{Data Collection Details} \label{subsec:data_collection_details}
\begin{figure}[htbp]
    \centering
    \includegraphics[width=\linewidth]{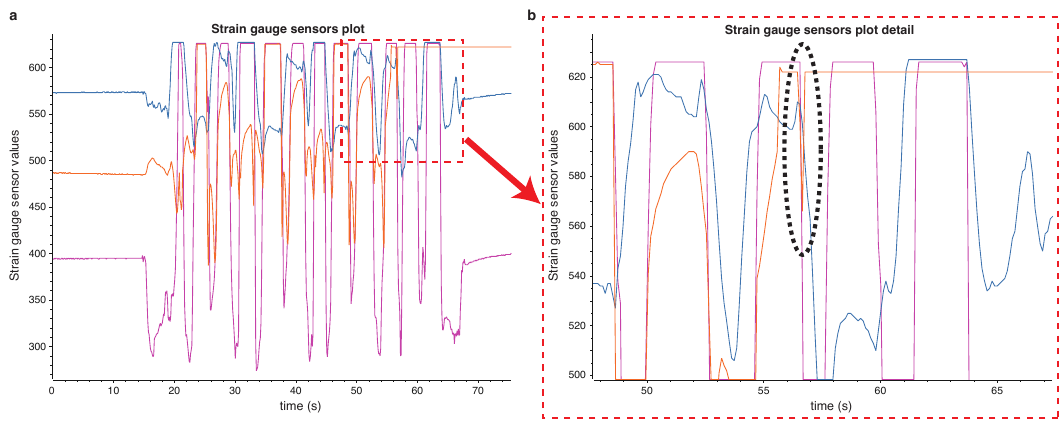}
    \caption{
      Identification of disconnected strain gauge sensors.
      \textbf{a,} Plot of strain gauge sensors, including those with disconnections. For clarity, a selected subset of strain gauge sensors is plotted.
      \textbf{b,} Enlarged view of a portion of (a). The strain gauge sensor highlighted in orange within the dashed region is observed to be disconnected.
    }
    \label{fig:supplementary-wire-disconnecting-plot}
  \end{figure}

{
  The true joint angle values were obtained using Visual SLAM \cite{Fuentes:2015:VisualSLAM} from Intel's RealSense T265 \cite{Schmidt:2019:RealsenseT265}, which provides the relative coordinates and pose of the joint from the initial position.
  The Intel RealSense T265 is commonly mounted on robots and is widely used for object tracking \cite{Bayer:2019:ExplorationHexapod}.
  Although Visual SLAM is less accurate than optical motion capture systems, its advantage is that it does not require expensive dedicated equipment, thereby enhancing the accessibility and reproducibility of this experimental framework for other researchers.
  In this study, the Intel RealSense T265 was attached to the bones forming the joint system.
  To obtain the relative coordinates and pose of the joint based on the camera's movement in its coordinate system from the initial position, a transformation matrix was used to convert the coordinate system.
  This transformation was performed using the TF package in the open-source Robot Operating System (ROS) \cite{Quigley:2009:ROS}.
  Regarding accuracy, comparisons with other cameras have been reported \cite{Chappellet:2021:VSLAMBenchmark}, and performance evaluations during the tracking of natural human head movements have also been conducted \cite{Hausamann:2021:EvaluationRealSenseT265}.
  In particular, \cite{Hausamann:2021:EvaluationRealSenseT265} reported that better accuracy is achieved during slower movements.
  When walking slowly over a distance of 80 m, the median absolute error for the yaw axis was found to be between 3 and 5 degrees, with a median relative error of 0.9 degrees.
  During data acquisition in this study, the joint was moved slowly, with the data collection period divided into smaller segments to minimize errors.
  To verify the reliability of the measurement system, the joint posture at the beginning and end of the data acquisition (\figref{fig:data-collection-for-joint-receptors}, time points 1 and 6) was compared.
  A relative offset was observed (X translation: \SI{0.001}{\meter}, Y translation: \SI{0.002}{\meter}, Z translation: \SI{0.004}{\meter}, Roll: \SI{0.203}{\degree}, Pitch: \SI{1.082}{\degree}, Yaw: \SI{1.053}{\degree}) during the measurement process.
  These correspond to approximately \SI{0.2}{\percent} of the full roll range (\SI{\pm 45}{\degree}) and \SI{0.6}{\percent} of the full pitch range (\SI{\pm 90}{\degree}), confirming that the offsets were sufficiently small compared with the total range of motion.
  Therefore, no additional correction was applied, and these offsets were treated as acceptable drift within the measurement tolerance.

  Since the fiber membrane made of latex is soft, it can stretch excessively when moving the joint, potentially applying stress to the lead wires of the strain gauge sensors and causing disconnection during the experiment (\figref{fig:supplementary-wire-disconnecting-plot}).
  In this study, we considered that damage to joint receptors is also a plausible occurrence in biological systems.
  Therefore, even if some receptors became disconnected during data collection, we deliberately included these data in the training set without applying any special processing.
}

\subsection*{Deep Learning Implementation} \label{subsec:deep_learning_implementation}
{
  It is well established that somatosensory information processing in the human body, particularly for deep sensory input, is transmitted via the dorsal column-medial lemniscus pathway.
  In this pathway, information from joint receptors is conveyed through first-order neurons in the dorsal root ganglia, second-order neurons in the medullary dorsal column nuclei, and third-order neurons in the thalamus before reaching the primary somatosensory cortex of the cerebrum.
  However, the specific mechanisms of information processing at each neuronal level remain unclear.
  In this study, deep learning, a technique inspired by neural network models in the brain, was used to approximate this sensory information processing.

  The deep learning architecture was designed as a four-layer fully connected network, modeled after the dorsal column–medial lemniscal pathway.
  While a time-series model could also have been adopted, a simpler feedforward structure was intentionally chosen to evaluate the potential extent to which proprioceptive information can be estimated directly from joint receptor signals, without assuming complex neural dynamics.
  The input layer, corresponding to the joint receptors, accepts 60 strain gauge sensor values.
  In the first layer (representing the primary neurons of the spinal dorsal root ganglion), the input is transformed into 128 dimensions (activated by a ReLU function).
  In the second layer (representing the secondary neurons of the medullary dorsal column nuclei), the 128 dimensions are reduced to 64 dimensions (activated by ReLU).
  The third layer (corresponding to the tertiary neurons in the thalamus) reduces the 64 dimensions to 32 dimensions (activated by ReLU).
  The output layer (representing the primary somatosensory cortex) converts the 32 dimensions into 6 output dimensions, corresponding to the joint position and orientation.
  The numbers of neurons in the hidden layers (128, 64, and 32) were chosen empirically, as is common practice in deep learning where the number of hidden units is treated as a hyperparameter to be determined experimentally depending on the task and dataset.
  Although these values were not derived from strict theoretical justification, we confirmed through experiments that this configuration provided sufficiently accurate performance in representing proprioception.
  Therefore, we retained this configuration in this study.

  The program was implemented in Python 3.8.10, using Pytorch 2.3.1 \cite{Paszke:2019:Pytorch} for deep learning.
  The loss function was MSELoss (mean squared error loss) implemented in Pytorch, and the optimization method employed was Adam \cite{Kingma:2014:Adam}.
  We used \SI{20}{\percent} of the collected data as test data and the remaining data as training data.
  The batch size was set to 32, with a learning rate of 0.001.
  The number of training epochs was set to 100 for both the experiment described in the Performance Verification of Joint Receptors Using Deep Learning section and for each loop in the Permutation Importance calculation in the Analysis of Redundancy in Joint Receptors section.
}

\subsection*{Proprioception Reproducibility Test} \label{subsec:proprioception_reproducibility_test}
\begin{figure}[htbp]
    \centering
    \includegraphics[width=\linewidth]{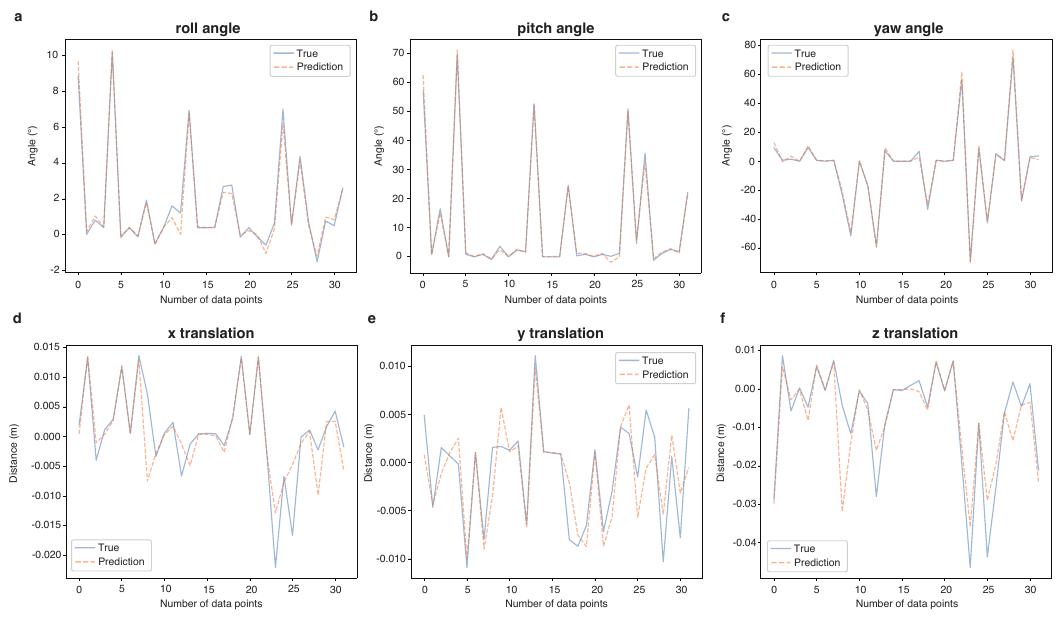}
    \caption{
      Validation of deep learning–based proprioception estimation with an additional dataset.
      This figure shows the performance of the trained model when tested on a dataset collected independently but with the same procedure as in the Performance Verification of Joint Receptors Using Deep Learning section.
      The biomimetic joint is an open-type ball joint loosely connected by soft tissues, which permits not only orientations but also translational movements. The predicted values are compared against actual measurements (ground truth).
      \textbf{a-f,}
      This graph illustrates the joint coordinates and orientation estimated using deep learning based on sensory receptor data obtained from 60 strain gauge sensors embedded in the joint capsule.
      The model was trained in the same manner as in the Performance Verification of Joint Receptors Using Deep Learning section but using a different dataset collected through the same method.
      Due to the open-type ball joint being loosely connected by soft tissues, the graph visualizes not only orientation(roll (a), pitch (b), and yaw (c)) but also translation(x (d), y (e), and z (f) positions), comparing them to actual measurements (ground truth).
      The vertical axis represents angles for roll, pitch, and yaw, as well as position in meters, while the horizontal axis represents a dimensionless number denoting the number of test data points.
      The results indicate a high prediction accuracy.
    }
    \label{fig:supplementary-joint-receptors-performance}
  \end{figure}

{
  To verify the reproducibility of proprioceptive estimation, data were re-acquired using the same method, and training was conducted again.
  Graphs verifying proprioception with the new model are presented in \figref{fig:supplementary-joint-receptors-performance}, and the statistical data are available in \tabref{tab:supplementary-proprioception-error-statistics-xyz} and \tabref{tab:supplementary-proprioception-error-statistics-rpy}.
  The results consistently demonstrated that proprioception could be achieved with an average error of less than 2 degrees for bending and twisting, confirming the reproducibility of the results.
}

\subsection*{Methods for Analyzing Redundancy and Receptor Importance in the Biomimetic Joint} \label{subsec:redundancy_analysis}
\begin{figure}[htbp]
    \centering
    \includegraphics[width=\linewidth]{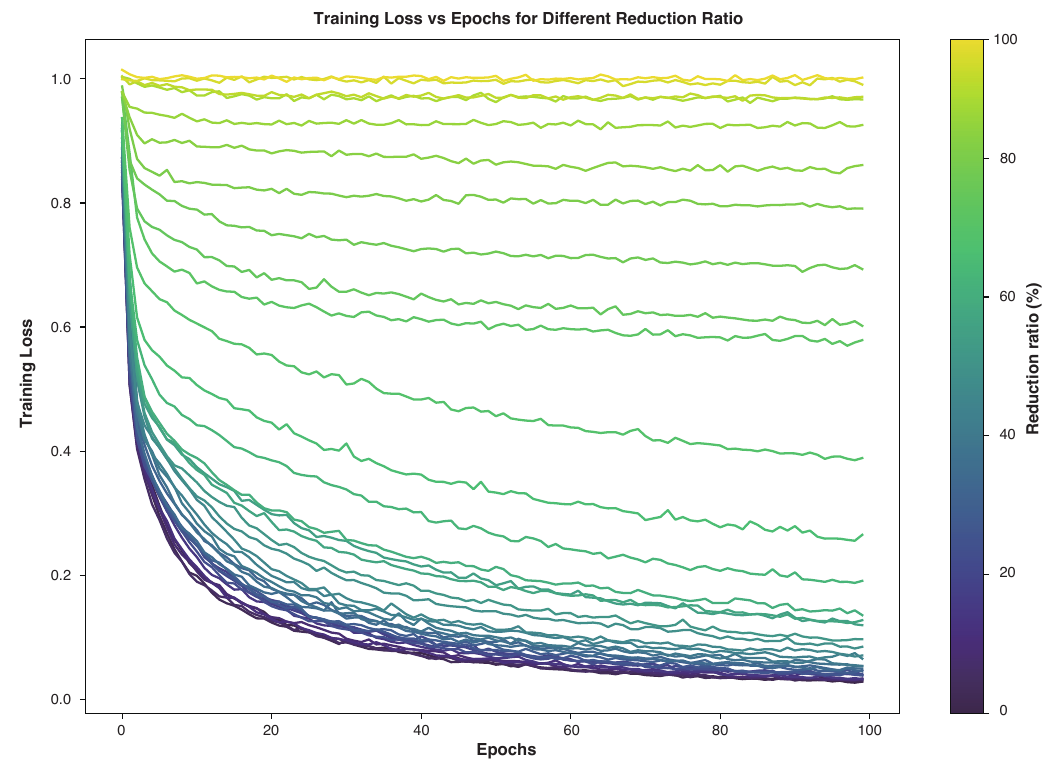}
    \caption{
      Training process for redundancy analysis.
      This graph illustrates the training process used to examine the redundancy of proprioception acquired through joint receptors, as discussed in the Analysis of Redundancy in Joint Receptors section.
      The horizontal axis represents the number of training iterations (epochs), while the vertical axis represents the loss function.
      The reduction ratio is represented by a continuous color gradient, as indicated by the color bar.
      As the reduction ratio increases, the training process exhibits slower convergence and higher final loss values.
    }
    \label{fig:supplementary-PI-loss-whole-all}
  \end{figure}

{
  For the analysis of redundancy in joint receptors described in the Analysis of Redundancy in Joint Receptors section, we employed Permutation Importance \cite{breiman:2001:RandomForests, altmann:2010:PermutationImportance, fisher:2019:PermutationImportance} to determine the order of sensor removal.
  The calculation of Permutation Importance was performed using the permutation\_importance function from the scikit-learn package (v0.22.2) \cite{Pedregosa:2011:ScikitLearn}, with 10 shuffling iterations.
  This method was used to evaluate the relative importance of each joint receptor to the model, and the most important receptor was iteratively removed in descending order of importance.
  The procedure was as follows.
  First, a deep learning model was trained using all joint receptor data, and its performance was evaluated.
  Next, the importance of each receptor to the model was assessed using Permutation Importance.
  Then, the most important receptor was removed, the model was retrained using the remaining receptors, and its performance was re-evaluated.
  This process was repeated until all receptors had been removed, and changes in model performance were analyzed throughout the reduction process.
  The convergence of the loss function during this process is shown in \figref{fig:supplementary-PI-loss-whole-all}, which illustrates that the removal of receptors made it increasingly difficult for the training to converge.
  To evaluate the significance of performance degradation under reduced receptor input, ten independent trials with different random seeds were conducted for each reduction ratio.
  Both mean and maximum angular errors were computed separately for pitch and roll.
  Statistical comparisons against the baseline condition (\SI{0}{\percent}) were performed using Welch’s t-test, with Holm correction applied to account for multiple comparisons.
  A threshold of $p < 0.05$ was used to determine statistical significance.

  For the analysis of receptor importance near the limits of joint motion, SHAP values \cite{lundberg:2017:SHAP} were employed, as they account for dependencies among features.
  Given the range of motion of the developed joint—up to \SI{45}{\degree} in twisting (roll angle) and up to \SI{90}{\degree} in bending (pitch and yaw angles)—we extracted subsets of data in which the roll angle exceeded \SI{20}{\degree}, the pitch or yaw angle exceeded \SI{70}{\degree}, or the absolute value of axial displacement exceeded \SI{2}{\milli\meter}.
  For the present analysis, SHAP values were aggregated across all translational and rotational components (x, y, z, roll, pitch, and yaw) to quantify the overall contribution of each receptor in near-limit conditions.
  To confirm the robustness of the results, we also performed an additional analysis using Permutation Importance.
  In the above range of conditions, the results obtained from SHAP and Permutation Importance were largely consistent.
  However, when further restricting the dataset to only the most extreme regions near the joint limits, Permutation Importance produced overly concentrated contributions toward a small number of receptors, and the trend did not align with the SHAP-based interpretation.
  These findings indicated that model training was unstable in this narrower region, and therefore those analyses were excluded.
  Each receptor was labeled as “adc(i)\_data(j),” where i denotes the board number and j denotes the sensor number.
  As shown in \figref{fig:biomimetic-joint}d, the sensors were arranged in three rows extending from the mid-capsular region toward the bone attachment region, which we referred to as the mid-capsular, transitional, and bone attachment regions.
  The SHAP analysis was conducted using the shap package v0.44.1 (Python module).
  The number of data points extracted under each condition was 415 for twisting (roll angles exceeding \SI{20}{\degree}), 207 for bending (pitch or yaw angles exceeding \SI{70}{\degree}), and 75 for axial displacements exceeding \SI{2}{\milli\meter}.
  Data collection near the joint limits was repeated carefully to avoid damaging the strain gauge sensors, as high loads could result in wire breakage, and convergence of the loss function during training was monitored to ensure data quality.
  The learning conditions were consistent with those used in the Performance Verification of Joint Receptors Using Deep Learning section.
  For interpretability, we focused on the top three receptors with the highest SHAP values in each trial.
  In the visualizations, receptors that appeared frequently among the top three across the five trials were highlighted in red, while the remaining top-ranked receptors were color-graded according to their SHAP values.
  All other receptors were displayed in dark gray (\figref{fig:joint-receptors-comparison-to-human}).
  Visualization was performed using Rhinoceros 3D Version 8 (Robert McNeel \& Associates, Seattle, WA, USA).
  Details of the feature analysis of the deep learning models are provided in the Supplementary Information.
}


\bibliography{sample}



\section*{Supplementary information}

\subsection*{Supplementary Methods: Feature Analysis Methods of Deep Learning Models}\label{subsec:feature_analysis_of_deep_learning_models}
\begin{figure}[htbp]
    \centering
    \includegraphics[width=\linewidth]{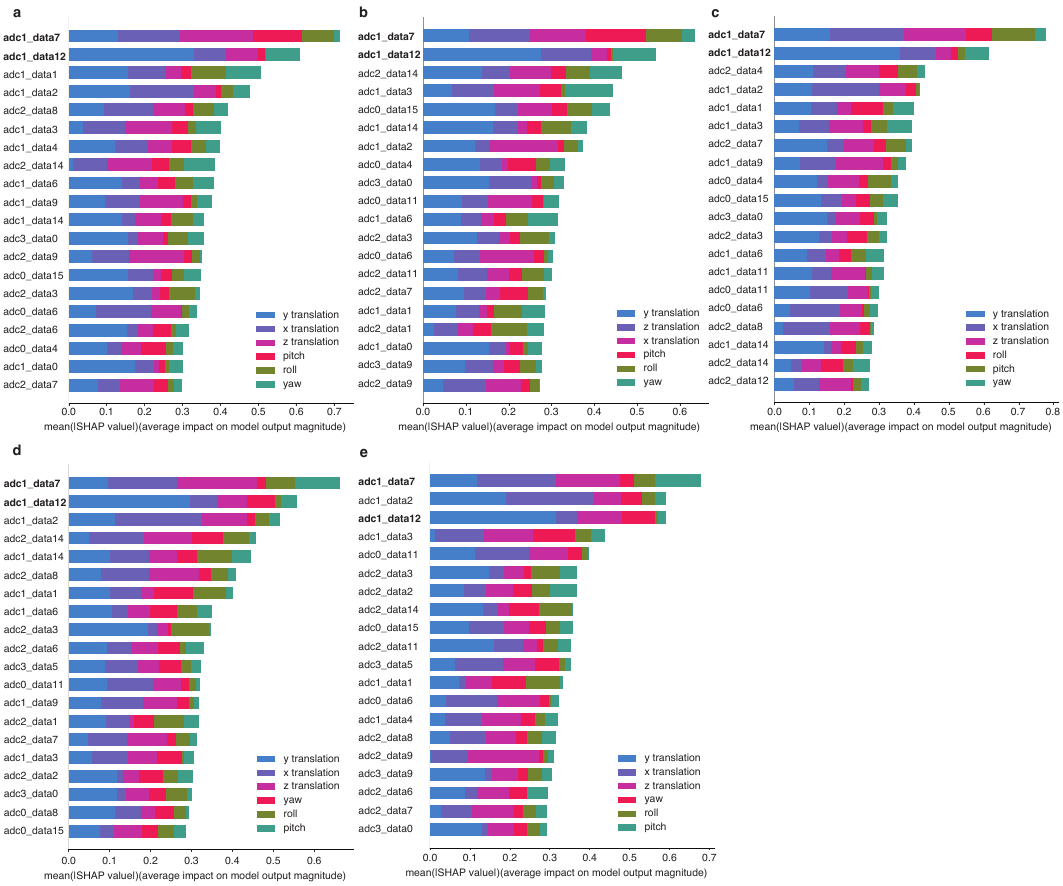}
    \caption{
      Distribution of critical joint receptors in near-limit regions where twisting(roll) angles exceeded \SI{20}{\degree}.
      \textbf{a-e,}
      Bar graphs showing the results of five trials conducted in regions where twisting(roll) angles exceeded \SI{20}{\degree} in the Distribution of Critical Sensory Receptors Near Range of Motion Limits section.
      For each trial, SHAP values representing the contributions of all motion components (x, y, z, roll, pitch, and yaw) were aggregated to evaluate the overall importance of each sensory receptor.
      Among the top three most important receptors in each trial, 3, 6, and 6 were located in the mid-capsular, transitional, and bone attachment regions, respectively, indicating that receptors situated at or near the bone attachment were predominant.
      Notably, adc1\_data7 and adc1\_data12 were consistently included among the top three across all trials.
      These receptors are located in the bone attachment and transitional regions, respectively.
    }
    \label{fig:supplementary-roll-feature-analysys}
  \end{figure}

\begin{figure}[htbp]
    \centering
    \includegraphics[width=\linewidth]{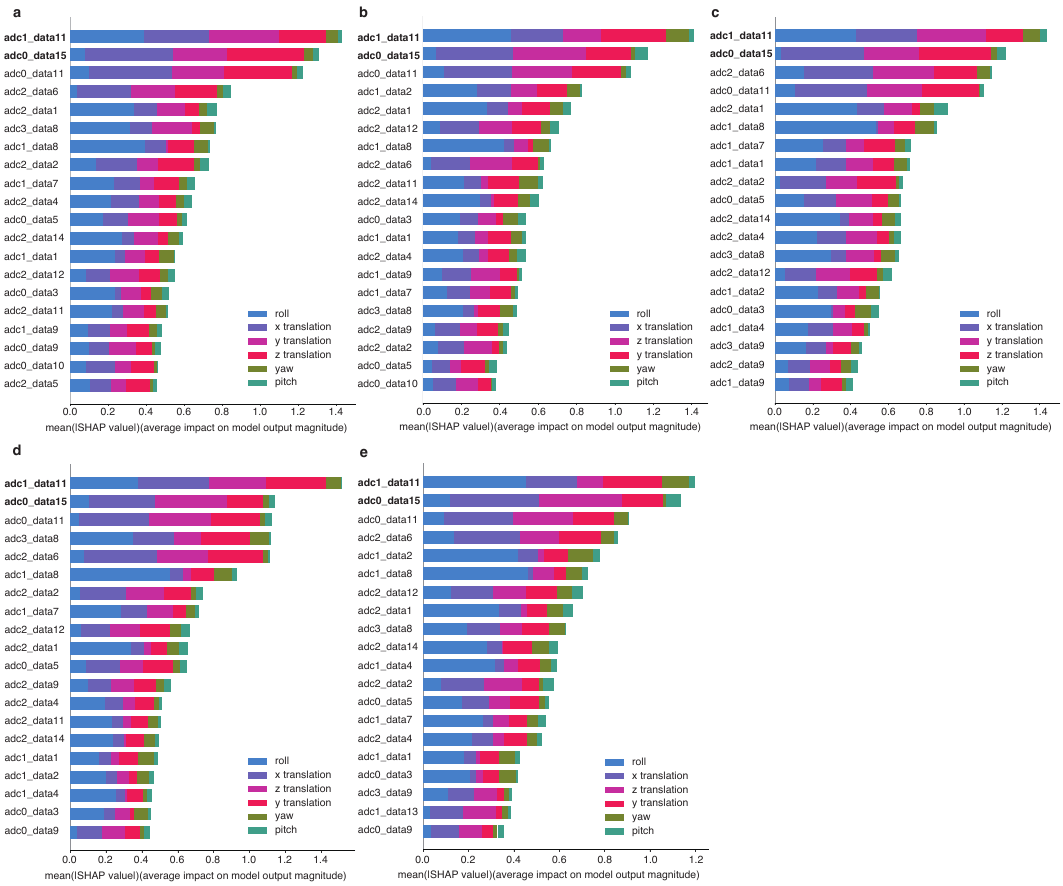}
    \caption{
      Distribution of critical joint receptors in near-limit regions where bending(pitch or yaw) angles exceeded \SI{70}{\degree}.
      \textbf{a–e,}
      Bar graphs showing the results of five trials conducted in regions where bending(pitch or yaw) angles exceeded \SI{70}{\degree} in the Distribution of Critical Sensory Receptors Near Range of Motion Limits section.
      For each trial, SHAP values representing the contributions of all motion components (x, y, z, roll, pitch, and yaw) were aggregated to evaluate the overall importance of each sensory receptor.
      Among the top three most important receptors in each trial, 0, 9, and 6 were located in the mid-capsular, transitional, and bone attachment regions, respectively, indicating that receptors situated at or near the bone attachment were predominant.
      Notably, adc0\_data15 and adc1\_data11 were consistently included among the top three across all trials.
      These receptors are located in the bone attachment and transitional regions, respectively.
    }
    \label{fig:supplementary-pitch-feature-analysys}
  \end{figure}

\begin{figure}[htbp]
    \centering
    \includegraphics[width=\linewidth]{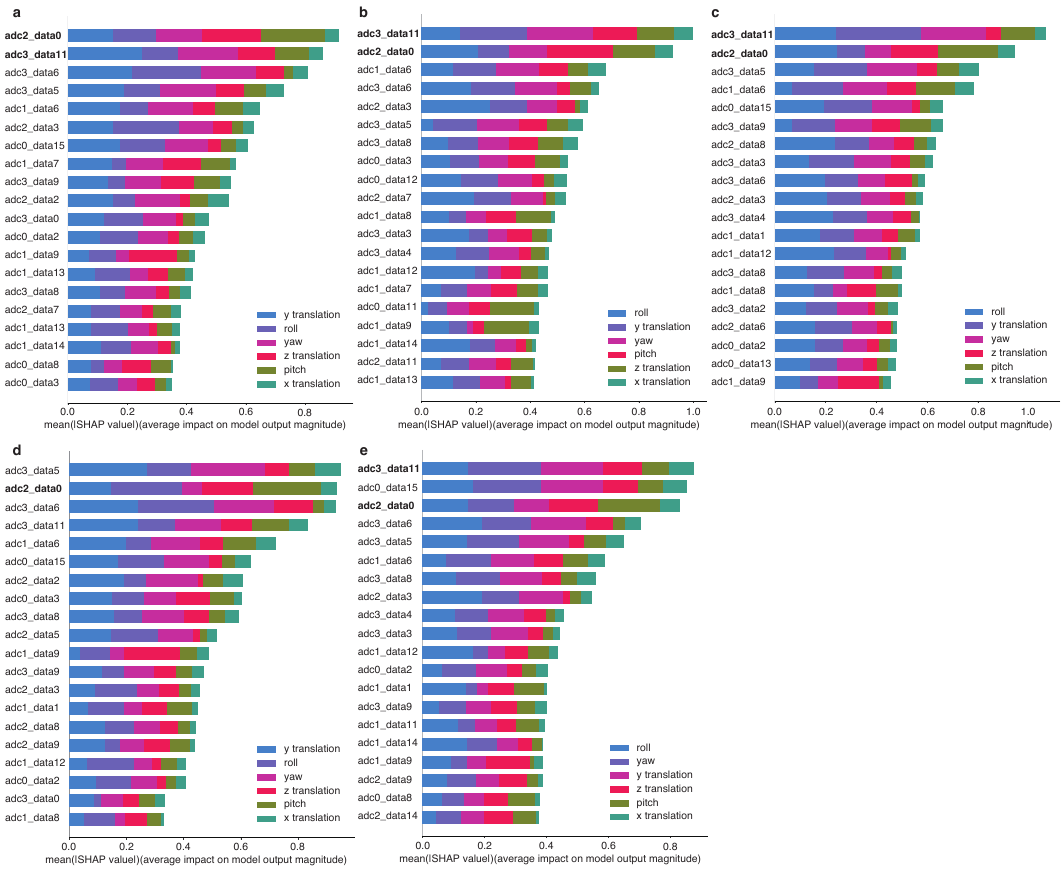}
    \caption{
      Distribution of critical joint receptors in near-limit regions where the absolute value of x-translation exceeded \SI{2}{\milli\meter}.
      \textbf{a–e,}
      Bar graphs showing the results of five trials conducted in regions where the absolute value of x-translation exceeded \SI{2}{\milli\meter} in the Distribution of Critical Sensory Receptors Near Range of Motion Limits section.
      For each trial, SHAP values representing the contributions of all motion components (x, y, z, roll, pitch, and yaw) were aggregated to evaluate the overall importance of each sensory receptor.
      Among the top three most important receptors in each trial, 5, 6, and 4 were located in the mid-capsular, transitional, and bone attachment regions, respectively, indicating that mid-capsular receptors were also frequently identified as critical.
      Notably, adc3\_data11 (transitional region) and adc2\_data0 (mid-capsular region) were frequently included among the top three across the trials.
    }
    \label{fig:supplementary-x-feature-analysys}
  \end{figure}

{
  Several methods have been proposed to quantitatively evaluate the importance of features in machine learning models, including Permutation Importance \cite{breiman:2001:RandomForests, altmann:2010:PermutationImportance, fisher:2019:PermutationImportance}, SHAP values \cite{lundberg:2017:SHAP}, and LIME \cite{ribeiro:2016:LIME}.
  Permutation Importance quantifies the importance of each feature by randomly shuffling it and measuring the resulting decrease in model performance.
  SHAP values, based on optimal transport theory, can capture interactions between features, but this method is computationally intensive. LIME, on the other hand, is specifically designed for explaining local predictions.
  In the redundancy analysis presented in the Analysis of Redundancy in Joint Receptors section, features were systematically reduced to examine the overall behavior of the model, requiring multiple iterations for each feature.
  To minimize computational costs while obtaining an overview of feature importance, Permutation Importance was chosen as the evaluation criterion.
  For the feature analysis in the Distribution of Critical Sensory Receptors Near Range of Motion Limits section, SHAP values were used despite their higher computational cost, as they account for dependencies between features from each joint receptor.
  In the Distribution of Critical Sensory Receptors Near Range of Motion Limits section, the model was trained five times for each of the twisting, bending, and push-pull movements, and the SHAP values were calculated.
  The results are presented in \figref{fig:supplementary-roll-feature-analysys} for the twisting movement, \figref{fig:supplementary-pitch-feature-analysys} for the bending movement, and \figref{fig:supplementary-x-feature-analysys} for the push-pull movement.
}



  \begin{table}[htbp]
    \centering
    \caption{Table of Statistics on Positional Errors in Different Datasets.}
    \small
    \begin{tabular}{|c|c|c|c|}
      \hline
      Error Type & x (m) & y (m) & z (m) \\
      \hline
      Maximum Error   & 0.0209 & 0.0230 & 0.0276 \\
      Mean Error      & 0.0021 & 0.0023 & 0.0031 \\
      Standard Deviation & 0.0033 & 0.0035 & 0.0047 \\
      \hline
    \end{tabular}
    \label{tab:supplementary-proprioception-error-statistics-xyz}
  \end{table}

  \begin{table}[htbp]
    \centering
    \caption{Table of Statistics on Attitude Errors in Different Datasets.}
    \small
    \begin{tabular}{|c|c|c|c|}
      \hline
      Error Type & roll (°) & pitch (°) & yaw (°) \\
      \hline
      Maximum Error   & 7.9132 & 5.3688 & 8.6223 \\
      Mean Error      & 0.3172 & 0.6780 & 1.2479 \\
      Standard Deviation & 0.7047 & 0.8202 & 1.6711 \\
      \hline
    \end{tabular}
    \label{tab:supplementary-proprioception-error-statistics-rpy}
  \end{table}

  \begin{figure}[htbp]
    \centering
    \includegraphics[width=\linewidth]{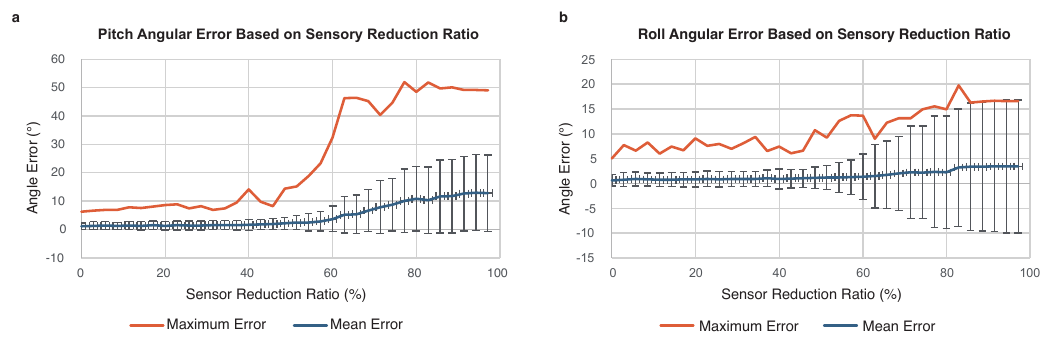}
    \caption{
      Redundancy analysis of proprioception estimation under reduced sensor input from a single trial.
      This figure illustrates the angular errors for pitch (a) and roll (b) in one trial.
      Error bars represent the within-trial standard deviation of errors without scaling.
      Because this is a single trial, the results are subject to stochastic variation inherent in deep learning, and no statistical analysis can be performed.
      Nonetheless, visual inspection suggests a noticeable degradation in performance around a reduction ratio of approximately \SI{50}{\percent}, which is consistent with the averaged results presented in \figref{fig:redundancy-of-joint-receptors}.
    }
    \label{fig:redundancy-of-joint-receptors-sample}
  \end{figure}

\section*{Funding}
  This work was supported by JSPS KAKENHI Grant Numbers JP24KJ0697.

\section*{Author contributions statement}
  A.Miki conceptualized the project and developed the methodology.
  A.Miki, S.Hasegawa,  S.Yuzaki, Y.Sahara, and R.Yoshimoto conducted the investigation.
  A.Miki handled visualization.
  A.Miki wrote the original draft.
  K.Kawaharazuka and K.Okada did the reviewing and editing of the manuscript.

\section*{Additional information}
\subsection*{Accession codes}  
  Data and code are available in https://github.com/poyotamu000/joint-capsule.
\subsection*{Competing interests}
  The authors declare no competing interests.






\end{document}